\documentclass[10pt,twocolumn,letterpaper]{article}
\usepackage[textwidth=6.875in,textheight=8.875in,left=0.8125in,top=1in,columnsep=0.3125in]{geometry}
\usepackage{times}
\usepackage[dvipsnames]{xcolor}
\usepackage{graphicx,amsmath,amssymb,booktabs,etoolbox,cuted}
\usepackage[numbers,sort&compress]{natbib}
\usepackage[hyphens]{url}
\usepackage[format=plain,labelformat=simple,labelsep=period,font=small,compatibility=false]{caption}
\usepackage[font=footnotesize,skip=3pt,subrefformat=parens]{subcaption}

\usepackage[shortlabels,inline]{enumitem}
\setlist[itemize]{noitemsep,leftmargin=*,topsep=0em}
\setlist[enumerate]{noitemsep,leftmargin=*,topsep=0em}
\font\subsectionheadingfont=ptmb scaled 1100
\font\subsubsectionheadingfont=ptmb scaled 1000
\makeatletter
\def\preprintsection{\@startsection{section}{1}{\z@}{-10pt plus -2pt minus -2pt}{7pt}{\large\bf}}
\def\preprintssect#1{\preprintsection*{#1}}
\def\preprintsect#1{\preprintsection{\texorpdfstring{\hskip -1em.~}{}#1}}
\def\section{\@ifstar\preprintssect\preprintsect}
\def\preprintsubsection{\@startsection{subsection}{2}{\z@}{-8pt plus -2pt minus -2pt}{5pt}{\subsectionheadingfont}}
\def\preprintssubsect#1{\preprintsubsection*{#1}}
\def\preprintsubsect#1{\preprintsubsection{\texorpdfstring{\hskip -1em.~}{}#1}}
\def\subsection{\@ifstar\preprintssubsect\preprintsubsect}
\def\preprintsubsubsection{\@startsection{subsubsection}{3}{\z@}{-6pt plus -2pt minus -2pt}{3pt}{\subsubsectionheadingfont}}
\def\preprintssubsubsect#1{\preprintsubsubsection*{#1}}
\def\preprintsubsubsect#1{\preprintsubsubsection{\texorpdfstring{\hskip -1em.~}{}#1}}
\def\subsubsection{\@ifstar\preprintssubsubsect\preprintsubsubsect}
\def\@maketitle{%
  \newpage\null\vskip .375in
  \begin{center}
    {\Large\bf\@title\par}
    \vspace*{12pt}{%
      \large\lineskip .5em
      \begin{tabular}[t]{c}\@author\end{tabular}\par
    }
    \vskip .5em
  \end{center}
}
\renewenvironment{abstract}{\centerline{\large\bf Abstract}\vspace*{12pt}\noindent\it\ignorespaces}{\vspace*{12pt}}
\makeatother
\date{}

\usepackage{cuted} %
\usepackage{caption} %

\usepackage{bm}
\usepackage{booktabs}
\usepackage{multirow}
\usepackage{colortbl}
\usepackage{xspace}
\usepackage{microtype}
\DeclareMathOperator*{\argmin}{arg\,min}
\DeclareMathOperator{\RMSE}{RMSE}
\DeclareMathOperator{\RMS}{RMS}

\newcommand{\xrel}{\tilde{x}}                 %

\newcommand{\hxrel}{\hat{\tilde{x}}}
\newcommand{\Rarc}{R_{\mathrm{arc}}}
\newcommand{\Hrel}{H_{\mathrm{rel}}}              %
\newcommand{\Ld}{L_d}
\newcommand{\E}{\mathbb{E}}
\newcommand{\loss}{\mathcal{L}}

\newcommand{\best}[1]{\textbf{#1}}
\newcommand{\second}[1]{\underline{#1}}

\newcommand{\down}{\ensuremath{\downarrow}}
\definecolor{oursrow}{HTML}{FBEFE6}           %

\definecolor{vidcell}{HTML}{DCEBF7}
\definecolor{eqcell}{HTML}{E6E6E6}

\makeatletter
\renewcommand\paragraph{\@startsection{paragraph}{4}{\z@}{0.9ex \@plus .1ex \@minus .1ex}{-1em}{\normalfont\normalsize\bfseries}}
\makeatother

\makeatletter
\newcommand{\paperdisplayskips}{%
  \setlength{\abovedisplayskip}{4pt \@plus 1pt \@minus 1pt}%
  \setlength{\belowdisplayskip}{4pt \@plus 1pt \@minus 1pt}%
  \setlength{\abovedisplayshortskip}{1pt \@plus 1pt}%
  \setlength{\belowdisplayshortskip}{3pt \@plus 1pt \@minus 1pt}}
\g@addto@macro\normalsize{\paperdisplayskips}
\makeatother
\paperdisplayskips

\definecolor{linkblue}{rgb}{0.12,0.30,0.52}
\usepackage[breaklinks,colorlinks,allcolors=linkblue]{hyperref}
\usepackage[capitalize]{cleveref}
\crefname{section}{Sec.}{Secs.}\Crefname{section}{Section}{Sections}
\crefname{table}{Tab.}{Tabs.}\Crefname{table}{Table}{Tables}
\title{Physical Laws without Physics Loss? Probing Learned Dynamics\\ in Video World Models through Forced Response}
\author{Yutao Li, Chen Wang\textsuperscript{*}, Wangmeng Zuo, Hui Li\textsuperscript{*}\\[2pt]
Harbin Institute of Technology\\[2pt]
{\small \textsuperscript{*}Corresponding authors: Chen Wang (\href{mailto:chen.wang@hit.edu.cn}{chen.wang@hit.edu.cn}), Hui Li (\href{mailto:lihui@hit.edu.cn}{lihui@hit.edu.cn})}}
\hypersetup{pdftitle={Physical Laws without Physics Loss? Probing Learned Dynamics in Video World Models through Forced Response},pdfauthor={Yutao Li, Chen Wang, Wangmeng Zuo, Hui Li},pdfsubject={Preprint}}

\usepackage{fancyhdr}
\fancypagestyle{plain}{%
  \fancyhf{}%
  \fancyhead[L]{\footnotesize\itshape\color{gray}Preprint}%
  \fancyfoot[C]{\thepage}%
}

\begin{document}

\maketitle
\begin{strip}
\centering
\includegraphics[width=\linewidth]{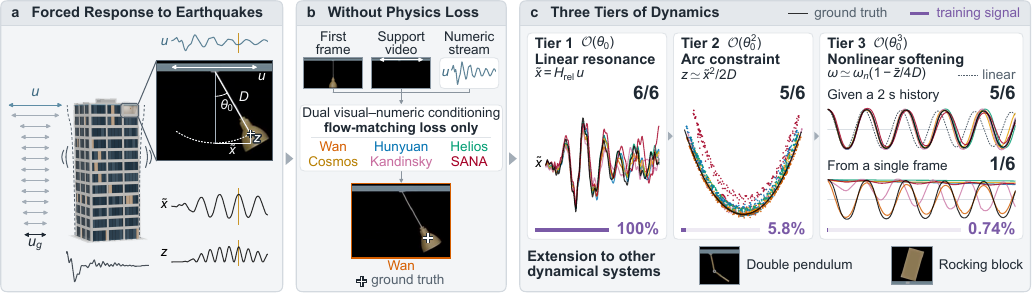}
\captionsetup{hypcap=false}
\captionof{figure}{\textbf{Physical laws are learned without a physics loss, tier by tier.}
(a) An earthquake shakes the ceiling holding a suspended object.
(b) Six backbones learn the response from the flow-matching loss alone.
(c) From Tier 1 to Tier 3 the training signal (bar: squared displacement in training, relative to the swing's) falls by two orders of magnitude, yet five or six backbones reproduce each tier.
Tier 3 separates law from state: released at 60$^\circ$, five follow the ground truth given a 2\,s history, only Wan from a single frame (three of four seeds).}
\captionsetup{hypcap=true}
\label{m:fig:teaser}%
\end{strip}

\begin{abstract}
Do video world models learn physical laws without a physics loss, or merely imitate motion?
Overall motion and appearance scores cannot settle it.
We investigate this through nonlinear forced response: an excitation shakes the support of a suspended object.
We introduce dual visual--numeric conditioning and adapt six image-to-video backbones to predict this motion using only the flow-matching loss.
Wan predicts motion accurately and generalizes to unseen excitations, objects, poses, and scenes.
Counterfactual tests suggest capabilities beyond replaying training videos.
But what laws have they learned?
We design three tiers of tests using Taylor expansions of the dynamics and geometry.
All six reproduce the linear response, including resonance; five capture higher-order dynamics, including nonlinear softening.
Remarkably, softening is barely present in training: only 0.74\% of the swing's squared displacement.
Extensions to a double pendulum and a rocking block show sustained dynamics are learned more faithfully than brief impacts.
Yet learned dynamics may fail to appear when state estimation or propagation goes wrong.
We turn these tests and findings into strategies to improve video training and generation.%
\end{abstract}
\section{Introduction}
\label{m:sec:intro}

Video generation models are increasingly cast as world models: given an image and a control signal, they simulate how a scene evolves~\cite{genie,vid2world,avid,cosmos,cosmos3,videozeroshot}.
Yet the physical world rarely evolves in isolation.
A robot moves objects on an accelerating base, and an earthquake shakes everything inside a room.
In such a \emph{forced response}, motion follows no preset visual script: it is the system's own dynamics answering an external excitation.
Does predicting pixels with standard generative training, and no physics loss, actually teach a video model the laws behind such motion, or only how motion looks~\cite{physlaw,physicsiq}?

Right now, we cannot tell.
Existing physical benchmarks typically follow a scene from an opening prompt or frames and summarize it with overall motion or appearance scores~\cite{physicsiq,physlaw,pisa,rigidbench}.
This leaves two blind spots.
First, under realistic shaking, most motion is a small-amplitude linear swing, which dominates the displacement and can hide whether a model violates the geometry of a rigid rod or the nonlinear dynamics.
Second, a single score conflates three abilities that mechanics keeps separate: \emph{reproducing a governing law}, \emph{inferring the state from pixels}, and \emph{propagating that state forward in time}.

When we separate these abilities, a striking paradox appears.
We adapt six image-to-video backbones to a suspended object under earthquake shaking, through one dual visual--numeric conditioning interface and only the flow-matching loss.
Among them, Hunyuan~\cite{hunyuan} is second only to Wan~\cite{wan} in trajectory error, and an appearance metric even ranks it above Wan.
Now release the object from rest at 60$^\circ$: Wan's object swings, while Hunyuan's stays frozen in place, as if gravity had vanished.
Yet show Hunyuan just 2\,s of a true swing ending in that very pose, and it swings at once, its period even lengthening with amplitude as a real pendulum's does (\cref{m:fig:teaser}c).
Stranger still, this nonlinear law is learned without a physics loss, although its terms hold only 0.74\% of the swing's squared displacement in training.
And Wan, trained only on rendered lamps, swings the lamp in a real photograph in time with the equation of motion.

How can a video model learn a higher-order law without a physics loss, yet fail to swing from a single frame?
To find out, we build a controlled forced-response laboratory in which the object's position is measured in millimeters.
The excitation reaches the object only through its support and no loss term states a law, so we call a law learned when the response obeys it on unseen cases, whatever its form inside the model.
The view is simple, yet the task is hard: before adaptation, no backbone produces a normal swing.
We probe three tiers, order by order: linear resonance at $\mathcal{O}(\theta_0)$, the arc constraint of a rigid rod at $\mathcal{O}(\theta_0^2)$, and nonlinear softening at $\mathcal{O}(\theta_0^3)$, alongside counterfactual interventions (\cref{m:fig:teaser}).

This mechanical lens shows that reproducing a law, inferring the state from a single frame, and passing it across windows can fail separately.
It explains how sampling and training choices shape a rollout, which passes its state from window to window like a numerical integrator.
Extended to a double pendulum and a rocking block~\cite{housner1963}, it also exposes brief impacts that rollouts miss.
Together, our findings turn ``do the pixels look plausible?'' into a sharper question: \emph{which laws are learned and up to what order, what state activates them, and what can a rollout carry forward?}
\section{Related Work}
\label{m:sec:related}

\paragraph{Evaluating physical fidelity in video models.}
Physical benchmarks have progressed from rating commonsense in text-to-video generation~\cite{videophy,phygenbench} and comparing continuations with real recordings~\cite{physicsiq,physicsiqverified} to testing quantitative mechanics: 2D laws~\cite{physlaw}, rigid-body motion apart from appearance~\cite{rigidbench}, dropping objects~\cite{pisa}, and equations of motion on real systems~\cite{morpheus}.
Across this progression, however, a score of the whole motion leaves open which orders of a law are learned and whether a failure lies in the law or in the inferred state.
We instead drive systems with strongly nonstationary excitations and probe each law order by order, using counterfactual edits and releases from rest to separate law, state, and propagation.

\paragraph{Physical control and physics-informed generation.}
Efforts to make video generators physically grounded follow two paradigms.
The first learns control interfaces from data, conditioned on trajectories~\cite{motionprompting,motionctrl}, actions~\cite{genie,avid,vid2world,cosmos,cosmos3}, a moving hand or robot~\cite{interdyn,acwmphys}, or forces, velocity changes, and goals~\cite{forceprompting,goalforce,streamforce,physstream}.
Generative Image Dynamics~\cite{gid} predicts oscillatory motion from one image without prescribed forcing.
The second injects physics explicitly, through simulators, property controls, physical priors, or foundation-model alignment~\cite{physgen,physdreamer,physctrl,physvid,phyco,wisa,videorepa,videojam}, or, as in physics-informed networks, through equation residuals~\cite{pinn} and built-in structure~\cite{hnn,phydnet}.
Where the first makes control the goal, we use it to test the laws beneath it.
Where the second adds a simulator or a physics loss to the generator, we add neither: a prescribed support excitation serves as a causal probe with an exact counterfactual ground truth.

\paragraph{Learned physics and state propagation.}
Self-supervised and generative models acquire structure they were never trained for, from segmentation~\cite{dino} and correspondence~\cite{dift} to board-game states~\cite{othello} and intuitive physics~\cite{intuitivephysics}.
Dynamical parameters can be estimated from video~\cite{learnphysvideo}, and learned simulators share one model across materials given particle states~\cite{gns}.
Yet how far a video model that sees only frames learns physical laws that no loss term states, and how this depends on each law's training signal, remains open.
Long rollouts must also pass the state across autoregressive windows, where the gap between oracle (ground-truth) training histories and the model's own lets errors compound, which Diffusion Forcing~\cite{diffusionforcing} and Self Forcing~\cite{selfforcing} address in training.
We complement such training by intervening on the history frames (true, frozen, or delayed) and on the stride, which shows what a window must inherit and how the stride, like a time step, changes the damping.
\section{A Controlled Forced-Response Laboratory}
\label{m:sec:testbed}

\begin{figure*}[t]
  \centering
  \includegraphics[width=\linewidth]{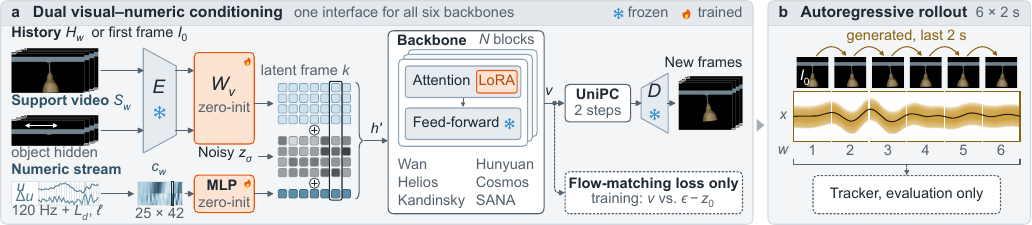}
  \caption{\textbf{One conditioning interface for all six backbones.} (a) The frozen encoder takes the history $H_w$ and the support video, and $c_w$ groups 20 samples each of $u$ and $\Delta u$ per latent frame with $\Ld$ and $\ell$, $25\times42$ for Wan; native channels or zero-initialized modules add both to the input tokens. (b) Each later window takes the last 2\,s of the backbone's own output as its history; below, a space-time slice of the generated 12\,s, ground truth in black.}
  \label{m:fig:interface}
\end{figure*}

\paragraph{Task.}
Given a first frame $I_0$ and a support excitation $u(t)$ sampled at 120\,Hz, we generate 12\,s of video at 24\,fps in which the object performs the forced response to $u$, with horizontal position $x(t)$ and vertical displacement $z(t)$.
The model also receives a fixed prompt and two numeric lengths, the equivalent length $\Ld$ and the rod length $\ell$, which specify the natural period.
A backbone could take the period from them or from the image, and wrong ones in \cref{m:sec:state} show that the backbones take it from the image.

\paragraph{Data.}
Excitations are floor responses of nonlinear building models under recorded earthquakes, cut to the strongest 12\,s, and the object starts at rest.
The view is chosen for measurement: a fixed orthographic side view on a black background, at 224$\times$160 pixels, removes perspective distortion and background clutter as causes of a missed law.
Each pixel spans 20.4\,mm, so the motion can be measured in millimeters.
Training uses 5,206 videos from 91 earthquake records and 15 buildings, each showing one suspended object at the center of the frame, a pendant lamp in one of eight shapes and five equivalent lengths (0.81--1.51\,m).
Testing uses 600 cases: 200 training videos as references and 400 unseen cases, each group changing one factor of training: new earthquake records, buildings, lengths (0.70--1.76\,m), or five objects other than lamps, such as a plant basket or a birdcage.
Excitations are strongly nonstationary: the strongest 2\,s of a 12\,s record hold a median of 32\% of its squared acceleration, against 17\% if uniform.

\paragraph{Equation of motion.}
The object and its rod form one rigid body about the moving pivot, whose angle $\theta$ follows
\begin{equation}
    \ddot\theta+2\zeta\omega_n\dot\theta+\omega_n^2\sin\theta=-\frac{\ddot u}{\Ld}\cos\theta,
    \label{m:eq:motion}
\end{equation}
where $\omega_n^2=g/\Ld$ and $\zeta=0.005$.
The equivalent length $\Ld=I_p/(md_c)=d_c+I_c/(md_c)$ sets the natural period through the mass distribution: mass $m$, pivot-to-center distance $d_c$, and moments of inertia $I_p$ about the pivot and $I_c$ about the center of mass.
A tracked point at distance $D$ from the pivot moves as $\xrel=x-u=D\sin\theta$ and $z=D(1-\cos\theta)$; integrating \cref{m:eq:motion} gives the ground truth.
Linearized, \cref{m:eq:motion} has an impulse response that decays over $1/(\zeta\omega_n)$, 58--78\,s for the training lengths, far beyond a 2\,s window.
A window's own excitation thus tells little about its swing, and each window must inherit its state from the last.
\section{Dual Visual--Numeric Conditioning}
\label{m:sec:adapt}

To give six backbones one conditioning interface, we add what each lacks: the excitation in two streams and autoregressive continuation (\cref{m:fig:interface}).
We adapt \mbox{Wan2.2-I2V-A14B}~\cite{wan}, \mbox{HunyuanVideo-1.5}~\cite{hunyuan}, \mbox{Helios-Base}~\cite{helios}, \mbox{Cosmos3-Nano}~\cite{cosmos3}, Kandinsky World Model~1.0~\cite{kandinsky}, and \mbox{SANA-Video}~2.0~\cite{sana}, called Wan, Hunyuan, Helios, Cosmos, Kandinsky, and SANA.
Five compress 24\,fps fourfold in time, a latent frame rate of $f_{\mathrm{lat}}=6$ per second, and SANA's LTX-2.3 VAE~\cite{ltx2,ltxvideo} eightfold.

\paragraph{Two streams of the excitation.}
The visual stream is a support video $S_w$ of the moving ceiling and pivot, with the object hidden and a mask $M_w$ of its known pixels, so it shows where the excitation acts without revealing the object's motion.
The numeric stream carries the 120\,Hz samples between the 24\,fps frames, which frames alone cannot resolve.
For each latent frame, a two-layer MLP maps the numeric condition $c_w$, made of the displacements and increments in that frame and the two numeric lengths, to one token $W_2\,\mathrm{SiLU}(W_1c_w)$.
This token is added to every token of the frame, so the numeric stream has no spatial index.
All six receive the same information through their native inputs wherever possible.
Each takes the history through its native conditioning, and Wan takes the support video the same way.
Hunyuan, Helios, Cosmos, and SANA also take the frozen encodings of history and support video through a projection $W_v$, and Kandinsky uses both pathways.
$W_v$ and $W_2$ start at zero, so training begins from the original backbone~\cite{controlnet}.

\paragraph{Training with the flow-matching loss alone.}
We freeze the base weights $\Theta_0$ and train only rank-16 LoRA~\cite{lora} updates on the attention layers and the zero-initialized modules, with flow matching~\cite{flowmatching}:
\begin{equation}
    \phi^*=\argmin_\phi\mathop{\E}\limits_{i,w,\sigma,\epsilon}\loss_{\mathrm{fg/bg}}\bigl(v_{\Theta_0,\phi}(a),\,\epsilon-z_0\bigr),
    \label{m:eq:objective}
\end{equation}
where $a=(z_\sigma,H_w,S_w,M_w,c_w,\sigma)$ holds the inputs, $z_\sigma=(1-\sigma)z_0+\sigma\epsilon$, $z_0$ is the target latent of window $w$ in case $i$, and $\loss_{\mathrm{fg/bg}}$ is a squared error weighting foreground and background by 0.7 and 0.3.
With this weight $W$, the minimizer over all functions is the weighted conditional expectation $v^*(a)=\E[W(\epsilon-z_0)\mid a]/\E[W\mid a]$.
Where the conditions determine the future, this minimizer follows the dynamics of the training videos, and \cref{m:sec:hierarchy} measures which tiers training reaches.
Training keeps image quality and consistency: a paired bootstrap test on eight VBench-I2V dimensions~\cite{vbench,vbenchpp} rules out a drop of 2.88\% or more in all six (\cref{m:tab:main}).
Only Hunyuan turns most of its moving videos still, a change these dimensions do not score.

\paragraph{Autoregressive rollout.}
The first window generates 2\,s from $I_0$.
Each later window re-encodes the last 2\,s of the backbone's own RGB output as its history $\hat H_w$ and adds 2\,s, a stride of 48 frames, so with $\hat H_1=I_0$ the rollout factorizes as \mbox{$p(\widehat V\mid I_0,u)=\prod_{w=1}^{6}p_{\Theta_0,\phi}(\widehat V_w\mid\hat H_w,S_w,M_w,c_w)$}.
Motion state passes between windows only through decoded frames, no window index enters, and training feeds oracle histories only (teacher forcing~\cite{scheduledsampling}).
As the conditions nearly determine a forced response, each window is sampled with two-step UniPC~\cite{unipc} at guidance scale 1, and training draws $\sigma$ from its two noise levels, 0.999 and 0.833 (\cref{m:sec:guidance}).
\section{Six Backbones Learn the Forced Response}
\label{m:sec:learn}

\begin{figure*}[t]
  \centering
  \includegraphics[width=\linewidth]{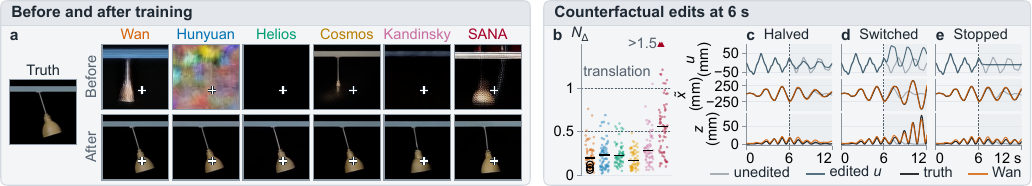}
  \caption{\textbf{All six backbones learn the forced response and follow edited excitations.} (a) An unseen case before and after training; $+$, ground truth. (b) $N_\Delta$ on 63 branches; bars, medians; circles, edits shown; dashes, thresholds. (c--e) Wan with $u$ halved, switched, or stopped.}
  \label{m:fig:qualitative}
\end{figure*}
\begin{table}[t]
  \centering
  \caption{\textbf{Motion and appearance scores do not show which laws hold.} Mean RMSE of $x$ and $z$ (mm) and median $Q_x$ on the 400 unseen cases, the VBench-I2V upper bound on the largest drop (\cref{m:sec:adapt}), and tiers reproduced (\cref{m:sec:hierarchy,m:sec:state}), Tier 3 given a 2\,s history or, as $3^\ast$, a single frame (Wan: three of four seeds).}
  \label{m:tab:main}
  \scriptsize
  \setlength{\tabcolsep}{3.0pt}
  \begin{tabular*}{\linewidth}{@{\extracolsep{\fill}}lccccccccc@{}}
    \toprule
    & \multicolumn{3}{c}{Motion} & Appear. & VBench-I2V & \multicolumn{4}{c}{Tier reproduced} \\
    \cmidrule(lr){2-4}\cmidrule(lr){5-5}\cmidrule(lr){6-6}\cmidrule(l){7-10}
    Backbone & $x$\,\down & $z$\,\down & $Q_x$\,\down & LPIPS\,\down & Drop (\%)\,\down & 1 & 2 & 3 & $3^\ast$ \\
    \midrule
Wan~\cite{wan} & \best{18.7} & \best{5.65} & \best{0.232} & 0.0262 & 0.94 & \checkmark & \checkmark & \checkmark & \checkmark \\
Hunyuan~\cite{hunyuan} & \second{22.3} & \second{6.86} & \second{0.262} & \second{0.0196} & 0.54 & \checkmark & \checkmark & \checkmark & $\times$ \\
Helios~\cite{helios} & 24.3 & 7.77 & 0.305 & 0.0249 & 0.96 & \checkmark & \checkmark & \checkmark & $\times$ \\
Cosmos~\cite{cosmos3} & 24.9 & 7.04 & 0.294 & \best{0.0187} & 0.87 & \checkmark & \checkmark & \checkmark & $\times$ \\
Kandinsky~\cite{kandinsky} & 25.7 & 9.18 & 0.319 & 0.0415 & 2.88 & \checkmark & \checkmark & \checkmark & $\times$ \\
SANA~\cite{sana} & 62.0 & 13.23 & 0.743 & 0.0369 & 1.98 & \checkmark & $\times$ & $\times$ & $\times$ \\
    \bottomrule
  \end{tabular*}
\end{table}

All six share data, 37,248 updates, and protocol, and we rank two backbones only when the 95\% bootstrap interval of their difference excludes 0.
Before training, no backbone produces a normal swing in any of 60 test cases, and most cannot even keep the object's shape (\cref{m:fig:qualitative}a).
We track $x$ and $z$ in every frame by template matching with ECC refinement~\cite{ecc}, and on the ground-truth videos of the main test its mean error stays below 1.7\,mm in $x$ and in $z$, under 0.1 pixels.
We report $\RMSE_x$, $\RMSE_z$, and $Q_x=\RMSE_x/\RMS[\xrel(t)-\xrel(0)]$, so an object fixed to the support scores $Q_x=1$.

\paragraph{The motion comes from the excitation.}
Editing the excitation is the sharpest test: an edit at 6\,s, unlike any training record, changes the motion despite the same history (\cref{m:fig:qualitative}b--e).
On 24 unseen cases, we halve, switch, or stop the excitation from 6\,s on, which gives 63 branches.
We score the generated change $\Delta\hxrel$ from the original branch against the true $\Delta\xrel$ by $N_\Delta=\RMS(\Delta\hxrel-\Delta\xrel)/\RMS(\Delta\xrel)$.
An object that only translates with the support scores 1, and a rollout that ignores the edit scores above 0.74 on every branch.
With $N_\Delta<0.5$ and a positive correlation as the rule, Wan follows 60 of 63 branches accurately, the other four 53--63, and SANA 27.

\paragraph{Appearance and motion can disagree.}
On the 400 unseen cases (\cref{m:tab:main}), Wan errs by 18.7\,mm in $x$, less than one pixel (20.4\,mm).
Over all 600 cases each of the other five errs more, Hunyuan least, by +2.73\,mm [1.12, 4.46], whereas a second training run of Wan differs by only $-0.4$\,mm, with similar probe results.
Appearance ranks the backbones differently: on crops aligned to the object, Cosmos and Hunyuan have lower LPIPS~\cite{lpips} than Wan, yet both err more in motion and misplace an off-center support (\cref{m:sec:guidance}).
Even when Wan's median $Q_x$ stays below 0.5 in all 13 conditions absent from training, other suspended objects among them, a low error only shows that the motion is close, not which laws it obeys.
\section{Five Backbones Learn All Three Tiers}
\label{m:sec:hierarchy}

\begin{figure*}[t]
  \centering
  \includegraphics[width=\linewidth]{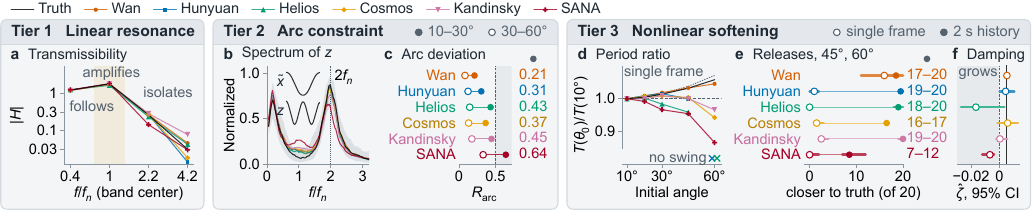}
  \caption{\textbf{Five backbones learn all three tiers; from a single frame, only Wan reproduces the softening, in three of four seeds.} (a) $|H|$ on the 550 cases tracked fully for all six. (b) Median $z$ spectra; gray, true quartiles. (c) Median $\Rarc$. (d) Single-frame period ratio; dotted, \cref{m:eq:period}; dashed, a linear pendulum; $\times$, no swing. (e) Releases closer to the ground truth than to a linear pendulum over four sampling seeds, median and range. (f) Estimated damping ratio; black, true videos.}
  \label{m:fig:hierarchy}
\end{figure*}
\begin{table}[t]
  \centering
  \caption{\textbf{One probe per tier, for three systems.} Each tier needs a more complete model than the one before; probes in parentheses.}
  \label{m:tab:probes}
  \scriptsize
  \setlength{\tabcolsep}{2.5pt}
  \begin{tabular}{@{}l>{\raggedright}p{2.3cm}>{\raggedright}p{2.4cm}>{\raggedright\arraybackslash}p{2.2cm}@{}}
    \toprule
    & Suspended object & Double pendulum & Rocking block \\
    \midrule
    Tier 1 & Linear resonance ($H$) & Two resonances, antiresonance ($|H|$) & Uplift onset (onset~time) \\
    Tier 2 & Arc constraint ($\mu$,~$b$,~$\Rarc$) & Arc of each rod ($\Rarc$) & Half-cycle grows with amplitude ($\eta_2$) \\
    Tier 3 & Nonlinear softening ($\eta$) & Softening of each mode~($\eta$) & Impact drop~\cite{housner1963} at 24\,fps ($\eta_3$) \\
    \bottomrule
  \end{tabular}
\end{table}

The motion builds up in three tiers, each one Taylor order higher in $\theta_0$ and each requiring a more complete model.
Tier 1 needs the linearized \cref{m:eq:motion}, Tier 2 adds the arc of the rigid rod, and Tier 3 the nonlinear terms of \cref{m:eq:motion}.
One probe per tier tells which of them a backbone reproduces (\cref{m:tab:probes}).
For Tiers 2 and 3, a backbone passes when its median probe value lies closer to the ground truth than to the lower-order approximation in every amplitude bin.

\paragraph{Tier 1: all six reproduce the linear resonance.}
At small angles \cref{m:eq:motion} filters the excitation into a swing of order $\theta_0$.
With $f_n=\omega_n/2\pi$ and $\Omega=f/f_n$, the ratio of absolute to support displacement at frequency $f$ is the transmissibility
\begin{equation}
    H(\Omega)=1+\frac{D}{\Ld}\,\frac{\Omega^2}{1-\Omega^2+2\mathrm{i}\zeta\Omega},
    \label{m:eq:transmissibility}
\end{equation}
whose second term $\Hrel(\Omega)$ gives the relative swing $\xrel=\Hrel u$.
The object follows the support for $\Omega\ll1$, resonates near $\Omega=1$, and for $D=\Ld$ is isolated above $\Omega=\sqrt2$.
Estimated in four bands of $\Omega$ from cross-spectra with the support, $|H|$ shows these three regimes for all six (\cref{m:fig:hierarchy}a), within 0.1--6\% of the ground truth at resonance (600 cases).

\paragraph{Tier 2: five satisfy the arc constraint.}
The excitation is horizontal, so $z$ can only come from the rigid rod, which keeps the object on a circle.
For a near-harmonic swing $\xrel=A\cos\varphi$,
\begin{equation}
    z=D-\sqrt{D^2-\xrel^{\,2}}\simeq\frac{\xrel^{\,2}}{2D}=\frac{A^2}{4D}\bigl(1+\cos2\varphi\bigr),
    \label{m:eq:arc}
\end{equation}
a slowly varying part and an oscillation at twice the swing frequency, with equal amplitudes.
This rise needs a swing-locked phase and an amplitude matched to the circle.

For the phase, the difference $\mu$ between the generated $z$ and the arc of \cref{m:eq:arc} computed from the backbone's own swing, in the $2f_n$ band, is 0 when the object rises at both ends of its swing and 180$^\circ$ when $z$ is inverted.
Power at $2f_n$ alone cannot show that $z$ is coupled to $\xrel$, so the cross-bicoherence $b\le1$ measures the quadratic phase coupling between the swing at $f_n$ and the rise at $2f_n$, against a surrogate level from random phases.
Every backbone's rise, which no input specifies, is phase-locked to its own swing near $2f_n$ (\cref{m:fig:hierarchy}b): the median $|\mu|$ against its own arc is below 4$^\circ$, and $b$ reaches 0.95--0.97, far above the surrogate level of 0.12.

For the amplitude, the arc deviation $\Rarc$ is the RMS distance of the generated object from its circle, relative to an object that keeps the true horizontal motion but never rises, so it is 0 on the circle and 1 for that object.
Five backbones keep it below 0.5 in both bins, whereas SANA reaches 0.64 at 10--30$^\circ$ (\cref{m:fig:hierarchy}c), so its rise violates the arc constraint.
All six track how the mean rise $A^2/(4D)$ varies, and Wan also matches its level.

\paragraph{Tier 3: five reproduce the softening.}
As $|\sin\theta|<|\theta|$, the restoring force softens and the undamped period grows with amplitude,
\begin{equation}
    T(\theta_0)=\frac{2T_n}{\pi}\,K\Bigl(\sin\frac{\theta_0}{2}\Bigr)=T_n\Bigl(1+\frac{\theta_0^2}{16}+\cdots\Bigr),
    \label{m:eq:period}
\end{equation}
where $T_n=2\pi/\omega_n$ is the natural period and $K$ is the complete elliptic integral of the first kind.
With the mean rise $\bar z=A^2/(4D)\simeq D\theta_0^2/4$ of \cref{m:eq:arc}, this reads $\omega\simeq\omega_n(1-\bar z/4D)$: the higher the object rises along the arc, the slower it swings.
Thus the period lengthens by 7.1\% from 10$^\circ$ to 60$^\circ$, which no linear pendulum can produce, and the phase lag $\theta_0^2\omega_nt/16$ shifts $\xrel$ at order $D\theta_0^3\omega_nt$.

An object on its circle has $\Rarc=0$ whatever its period, so we release it from $\pm$10--60$^\circ$ without excitation.
No excitation then imposes a period, so it must come from the backbone's own dynamics, and we estimate $\widehat T$ with a Lomb--Scargle periodogram~\cite{lomb,scargle} on a frequency grid far finer than FFT bins.
With $R_T=\widehat T(\theta_0)/\widehat T(10^\circ)$ at the same length and direction, the reproduced fraction $\eta=(R_T^{\mathrm{gen}}-1)/(R_T^{\mathrm{true}}-1)$ separates the amplitude dependence from any error in a backbone's natural period: $\eta=1$ is exact, and $\eta=0$ is a linear pendulum.

After a 2\,s history of a true swing ending in the release pose, all six keep swinging.
Five then reproduce the softening at both angles in every sampling seed, with a median $\eta$ of 0.60--1.32 and 16--20 of the 20 releases closer to the ground truth, whereas SANA does so only at 45$^\circ$ (\cref{m:fig:hierarchy}e).
To rule out a copied period, we replace the history with a linear swing ending in the same frame, whose period ignores amplitude.
Copying would remove the softening, yet $\eta$ falls by only 0.02--0.20, and this drop fades within 6\,s: the state at release mainly sets the period.
The softening also appears under excitation: on windows swinging to 25--45$^\circ$ from oracle histories, all six have a median $\eta$ above 0.5.

\begin{figure*}[t]
  \centering
  \includegraphics[width=\linewidth]{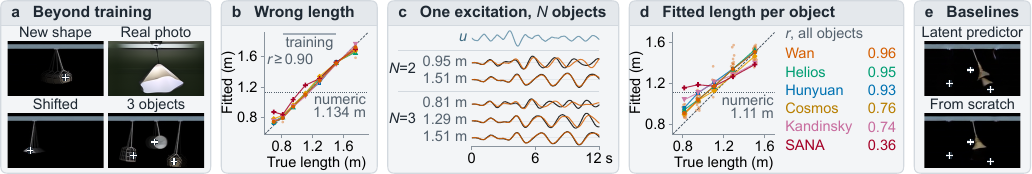}
  \caption{\textbf{Beyond training, the natural period comes from the image.} (a) Wan beyond training and on a real photograph; $+$, ground truth. (b) Fitted length under one wrong numeric length (dotted); dots, Wan; lines, all six. (c) $\xrel$, two scenes; orange, Wan; black, ground truth. (d) As (b), per object in such scenes. (e) Two baselines, scene of (a).}
  \label{m:fig:general}
\end{figure*}

\paragraph{The tier with the weakest training signal is still learned.}
In squared displacement the tiers enter at $\theta_0^2$, $\theta_0^4$, and $\theta_0^6$, so when most swings are small, each holds a smaller share than the one before: over 2\,s training windows restarted from the true state, the rise and the nonlinear correction amount to 5.8\% and 0.74\% of the squared displacement of the swing.
Weighted as in \cref{m:eq:objective}, the latent targets of the five VAEs keep this order, and the nonlinear term is rare as well as weak, held mostly by the 1.0\% of windows that swing beyond 45$^\circ$.
Tier 3 thus leaves the weakest training signal, yet five reproduce it once a 2\,s history supplies the state.
Take the history away, and they split.

\section{A Law Needs a State}
\label{m:sec:state}

A law needs an initial state and a natural period, and a backbone must infer both from its inputs.
We take the state first, then the period.

\paragraph{Only Wan reproduces the softening from a single frame.}
Why, then, does Hunyuan's object stay put when released at 60$^\circ$?
A second-order system needs a position and a velocity.
A 2\,s history shows both, whereas a single frame shows only a pose, so the backbone must initialize a resting state at an unseen pose.
Wan lengthens its period as a real pendulum does, with a median $\eta$ above 0.5 at both angles in three of the four seeds.
In 11--18 of its 20 releases from 45$^\circ$ and 60$^\circ$, the period is closer to the ground truth than to a linear pendulum (\cref{m:fig:hierarchy}d,e).
The others manage this in at most 4: they shrink the swing, keep a constant period, shorten it, or do not swing at all, as for Hunyuan and Helios in all 40 releases from 60$^\circ$.
From 10$^\circ$ and 20$^\circ$, near the training videos' rest pose, all six swing in at least 78 of the 80 releases.

Damping splits them only from a single frame.
Once real excitation stops, all six decay like a real swing, by 2--6\% per window against 3\%.
From a single frame, Wan's estimated damping ratio is 0.0053 against 0.005 (\cref{m:fig:hierarchy}f), a constant in no input or loss term, whereas the swings of Kandinsky, Helios, and SANA mostly grow, which no passive object allows.

\paragraph{The natural period comes from the image.}
Beyond position and velocity, a law also needs the natural period.
To test where a backbone gets it, we fit \cref{m:eq:motion} under the known excitation to its horizontal motion, with a free amplitude scale, and call the best-fitting length the fitted length $\hat L$, which sets the natural period.
The numeric lengths and the first frame could both set the period, and training never separates them, as each shape and rod has fixed numeric lengths, so we put them in conflict.
On one object, switching from the short to the long numeric lengths changes their natural period by 17\% but that of Wan's fitted length by a median of 0.31\%.
When every case receives one numeric length of 1.134\,m while the true $\Ld$ spans 0.70--1.76\,m, the fitted length still tracks the true one at $r=0.90$--$0.99$ (\cref{m:fig:general}b).
For four backbones, it also follows the whole body's mass distribution, as a compound pendulum does.
Regressed on the offset $D/\Ld-1$ of a point mass at the tracked point, their relative error $\hat L/\Ld-1$ has slopes of $-0.03$ to $0.02$, where the compound pendulum predicts 0 and the point mass 1.

The equation of motion, driven by the support motion observed at 24\,fps, instead uses the numeric length.
Given the true length, it errs by 11.5\,mm, yet given a wrong one its error rises to 175--214\,mm, whereas Wan stays at 38--39\,mm.
The equation integrates the length it is given, whereas the backbones infer it from the image.

Training shows only one object, whereas here $N=2$ or 3 objects of unseen shapes hang under one excitation, each with its own equivalent length (\cref{m:fig:general}a).
The numeric stream carries one set of lengths with no spatial index, so a model taking its period from it would swing all objects alike.
Each instead swings at its own period (\cref{m:fig:general}c), and over 120 objects and four seeds, the fitted lengths of Wan, Helios, and Hunyuan track the true ones at $r=0.92$--$0.96$ (\cref{m:fig:general}d).

These fitted lengths extrapolate.
At 0.70\,m and 1.76\,m, those of Wan, Hunyuan, and Helios lie beyond the training range on the true length's side in every case, wrong numeric length, and seed, and those of Cosmos in at least 33 of 36.
Wan's median error is at most 6.3\%, with the nearest training lengths 14--17\% away, and no median of the four falls back to them, as the case-based imitation of training examples reported for video models would~\cite{physlaw}.

\begin{figure*}[t]
  \centering
  \includegraphics[width=\linewidth]{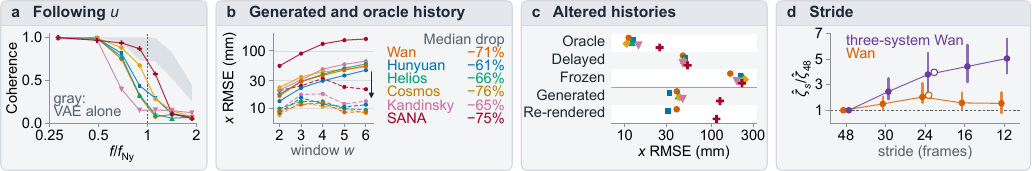}
  \caption{\textbf{Mechanics guides generation: what a window can follow and must inherit.} (a) Pivot coherence with $u$ against $f/f_{\mathrm{Ny}}$; gray, range of the VAEs alone. (b) $x$ RMSE per window, generated (solid) or oracle (dashed) history; key, colors of (a--c) and median drop. (c) The oracle history delayed or frozen, and the generated one re-rendered. (d) Damping ratio relative to stride 48, 95\% intervals; open, a second seed.}
  \label{m:fig:diagnosis}
\end{figure*}

\paragraph{A closer fit need not extrapolate.}
To separate fitting from extrapolation, we train two baselines on the same data with Wan's frozen VAE but no pretrained generative backbone: a deterministic latent predictor and a small Wan from scratch with as many trainable parameters as adapted Wan.
On 120 validation cases they err only 0.46--0.56 times as much as Wan, and given a 2\,s history both pass the softening probe.
Yet from a single frame at 45$^\circ$ or 60$^\circ$, neither keeps the object's shape, and their fitted lengths extrapolate in only 14--18 of the 36 fits, against all 36 for Wan.
With three objects, both collapse the scene into one central object (\cref{m:fig:general}e), the layout of every training video, whereas all six backbones keep every object.
Fitting these videos teaches the softening given a history, yet only the pretrained backbones reach poses and scenes that training never showed.

\paragraph{The dynamics transfer to a real photograph.}
From a photograph of a real lamp, with no retraining, Wan's swing keeps the timing of \cref{m:eq:motion} under six unseen records, at fitted lengths of 1.44--1.63\,m against 1.41--1.57\,m from its geometry at 20.4\,mm per pixel.
It hangs nearly still without excitation, and the natural period again comes from the image (\cref{m:fig:general}a).

\section{Strategies for Training and Generation}
\label{m:sec:guidance}

As the backbones reproduce the dynamics, mechanics explains each choice, within a window and across windows.

\paragraph{Deterministic dynamics favor regression over sampling.}
Single-step prediction already serves depth and normal estimation~\cite{garcia2025finetuning,lotus}, and one error budget explains why two sampling steps, guidance scale 1, and learning-rate annealing suit deterministic video dynamics as well.
Let $\Lambda$ be the covariance of $z_0$ given the conditions $c=(H_w,S_w,M_w,c_w)$; as \cref{m:eq:motion} determines the future, it holds only what $c$ leaves open, such as the sub-pixel state.
Then $N$ steps at guidance scale $\gamma$ give, up to cross terms and without the foreground weight,
\begin{equation}
    \E\|\hat z-z_0\|^2\simeq(1+\rho_N)\,\mathrm{tr}\,\Lambda+(\gamma-1)^2D_\gamma^2+\mathcal{E}_\phi,
    \label{m:eq:budget}
\end{equation}
where $\rho_N$ is the share of $\mathrm{tr}\,\Lambda$ that sampling adds, $D_\gamma$ the gap between guidance branches, and $\mathcal{E}_\phi$ the excess of trained weights over the minimizer.

For an ideal minimizer, one step returns the first guess $\E[z_0\mid c]$, so $\rho_1=0$, and an exact sampler draws from the posterior, so $\rho_\infty=1$~\cite{perceptiondistortion}.
Two-step UniPC adds only scaled initial noise to the mean, which gives $\rho_2\le1/4$ for a Gaussian posterior.
In practice, on a Wan checkpoint trained at six noise levels, one to three steps reach squared $x$ errors of 1.1--1.5\,$\hat\Lambda$ and four to 40 steps 2.0--2.4\,$\hat\Lambda$, with $\hat\Lambda$ the counterpart of $\mathrm{tr}\,\Lambda$ in $x$, fitted to its sampling path.
Guidance~\cite{cfg} only adds the disagreement of its two branches, which agree at the minimizer because half of the updates train the negative-prompt branch on the same history, excitation, and target.
Wan's squared error grows linearly in $(\gamma-1)^2$ over four scales, and scale 3.5 improves no backbone in validation.

Annealing shrinks $\mathcal{E}_\phi$, as a constant learning rate leaves the weights fluctuating~\cite{sgdbayes}: over 4,000 updates it cuts the error of four backbones by 46--56\%, and the same updates at a constant rate leave Wan at 25.8\,mm, against 18.7\,mm with annealing.
For deterministic dynamics, the budget thus favors few steps, guidance scale 1, and annealing.

\paragraph{A window follows the excitation up to about the latent Nyquist frequency.}
A window sees the excitation through its latent frames, whose Nyquist frequency is $f_{\mathrm{Ny}}=f_{\mathrm{lat}}/2$.
The generated pivot of every backbone keeps a coherence with $u$ of at least 0.96 below $0.6f_{\mathrm{Ny}}$ and at most 0.15 above $1.25f_{\mathrm{Ny}}$ (\cref{m:fig:diagnosis}a), whereas the VAEs alone keep this band: at 1.8--2.4\,Hz, the LTX-2.3 VAE keeps 0.88 and SANA 0.05.
From an oracle history, Hunyuan and Cosmos match Wan despite coarser spatial latents, whereas SANA, at half their latent frame rate, errs 2.47 times as much per window.
Here the band above the cutoff holds under 0.4\% of the displacement power of $u$, and \cref{m:eq:transmissibility} isolates the object from it.
In these backbones, the fastest excitation that a window follows thus scales with the latent frame rate.

\paragraph{The conditioning pathway shapes how position is used.}
The excitation acts at the pivot, and the numeric stream has no spatial index, so the pivot's position reaches a backbone only through the frames it is given.
When the support is shifted 750\,mm from the center where every training video places it, Wan and Kandinsky follow it, whereas the other four are biased toward the center by medians of 75--288\,mm, mostly because the generated pivot itself slides.
Given a native channel for its support video, Hunyuan follows the shift, as Wan does through either pathway, so off-center scenes should test any new interface.

\paragraph{A window must inherit position and velocity.}
Turning from a single window to the full rollout, a second-order system needs both position and velocity at every boundary, yet the rollout passes its state only through decoded RGB frames.
Linearized around the true state, its error follows $e_{w+1}\simeq J_we_w+\xi_w$, and light damping shrinks it only by $e^{-\zeta\omega_n\tau}=0.96$--$0.98$ over $\tau=2$\,s.
Replacing the history frames with true ones removes the inherited term $J_we_w$: on the 24 unseen cases of \cref{m:sec:learn}, this oracle history removes 61--76\% of the horizontal error after 2\,s (median per case; \cref{m:fig:diagnosis}b).
Each window adds about one oracle-history error $\xi_w$, including a lag of 3--5\% of a latent frame that the next window inherits almost fully.
The mean error therefore grows almost linearly in $w$, whereas independent errors would grow as $\sqrt w$.
Reproducing the dynamics and passing the state are thus separate targets for training.

\paragraph{The history must carry motion and phase.}
An oracle history corrects position, phase, amplitude, and appearance at once, so we alter it, or re-render a rollout, to see what a window relies on (\cref{m:fig:diagnosis}c).
Frozen at its last frame, the oracle history keeps pose and appearance but no motion, and every backbone errs an order of magnitude more: the backbones use the motion in their history.
Delayed by two frames, the oracle history already makes five backbones err more than their own rollouts.
Re-rendering a rollout's own swing as clean frames leaves its error almost unchanged, so a rollout can carry an inaccurate state even through clean frames.
Training sees only oracle histories, never phase-shifted ones, which points to training on a rollout's own~\cite{diffusionforcing,selfforcing}.

\paragraph{A rollout has a time step.}
Beyond what the history contains, how often windows advance also matters: we change only the stride, the number of the 48 new frames that a window keeps before the next one starts.
Keeping fewer raises the estimated damping ratio after a 2\,s history, up to about twofold for Wan and fivefold for the three-system Wan of \cref{m:sec:systems}, while the softening passes at every stride (\cref{m:fig:diagnosis}d).
Like a numerical integrator, whose damping depends on its time step~\cite{hht}, a rollout should be checked across strides.
\section{Extension to Other Dynamical Systems}
\label{m:sec:systems}

\begin{figure}[t]
  \centering
  \includegraphics[width=\linewidth]{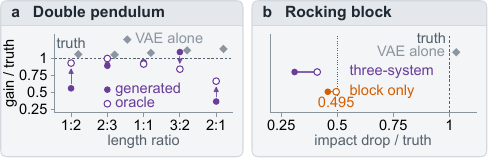}
  \caption{\textbf{The findings extend to other dynamical systems.} (a) Second-mode peak gain. (b) Drop in angular speed at an impact; dotted, Tier-3 criterion. Filled, generated history; open, oracle.}
  \label{m:fig:systems}
\end{figure}

Everything so far rests on one kind of system, a suspended object.
Are laws still learned, do they still need a state, and do they still depend on what a window inherits when the mechanics differ in kind?
We add a double pendulum, whose two modes couple, and a rocking block~\cite{housner1963}, which lifts off its floor and impacts it, as furniture does in an earthquake (\cref{m:fig:teaser}c).
One Wan learns all three through the same interface and loss, as a world model of a shaken room must: the systems share rank-16 LoRA adapters, and a system label selects small per-system adapters.
Each system's tiers follow its own hierarchy of models (\cref{m:tab:probes}), which for the block adds statics, rocking, and the impact in turn.

\paragraph{Laws of other kinds are learned as well.}
The double pendulum's tip errs by 25.5\,mm, against 121.6\,mm for rods fixed to the support, and both resonances and the hinge antiresonance lie within one frequency step of the sine drives at all five length ratios.
Both rods stay on their arcs at 30--60$^\circ$, and the first mode softens.
The model times the block's uplift, and the block rocks with a half-cycle that lengthens with amplitude ($\eta_2=0.72$).
Sharing one model keeps the object's three tiers.

\paragraph{A law again needs a state.}
From a single frame, the double pendulum's second mode stops within about 2\,s, whereas after a 2\,s history it keeps swinging and softens at 45$^\circ$, the split that \cref{m:sec:state} found for the object.

\paragraph{What a window inherits again shapes the dynamics.}
The oracle history lowers the error of all three systems: by 67\% on the object, within the six backbones' range, by half at the double pendulum's tip, and by about a fifth on the block.
It brings the double pendulum's second-mode peak gain to 0.67--1.00 of the ground truth's, so Tier~1 passes at all five length ratios (\cref{m:fig:systems}a).

\paragraph{The tiers find what rollouts still miss: an impact.}
Through the VAE alone, the ground-truth videos of all three systems keep every tier, including the drop in angular speed at an impact ($\eta_3=1.03$).
In the rollouts, this drop reaches 0.31 of the ground truth's, and over the whole rollout neither more sampling steps nor the oracle history lifts it to half (\cref{m:fig:systems}b).
The softening acts on every cycle, whereas an impact is brief, a possible reason why the latents hold it and rollouts do not yet, so brief events are a target for training.

\paragraph{A trajectory error can hide a law either way.}
The oracle history lowers the block's error yet leaves the impact short, so a lower error can hide a failed tier.
Removing the small residual adapters at inference keeps the object's three tiers while its error grows 4.5-fold, mostly because its fitted length falls to 0.90 of the true one, so a higher error can hide a preserved law.
Accuracy and a law are thus separate targets, and the tiers tell them apart.
\section{Conclusion}
\label{m:sec:conclusion}

Does predicting pixels, with no physics loss, teach a video model physical laws, or only how motion looks?
In our tests, the backbones learn to reproduce these laws to varying degrees.
Given a 2\,s history, five of six reproduce all three tiers, even a softening that training barely shows.
With the same history, two models trained from scratch also learn this softening and fit the data more closely.
Yet they fail on unseen poses and scenes the pretrained backbones reach.

Why, then, can a learned law fail to appear?
A law alone moves nothing.
Like an equation of motion, it needs an initial state, and a rollout must pass that state from window to window.
Progress on real footage likely needs a stronger training signal for brief events such as impacts.
It also needs models that infer the state from a single frame and pass it on like a faithful numerical integrator.
After all, a wrong video need not mean a missing law.
It may mean the wrong state.
{
    \small
    \clearpage
\bibliographystyle{references}   %
    \bibliography{main}
}

\end{document}